\documentclass[11pt,a4paper]{article}
\usepackage[hyperref]{acl2020}
\usepackage{times}
\usepackage{latexsym}
\usepackage{CJKutf8}
\usepackage{adjustbox}
\usepackage{graphicx}
\usepackage{xcolor}
\usepackage{booktabs}
\usepackage{algorithm2e}
\usepackage{amsmath}
\usepackage{amsfonts}
\usepackage{booktabs}
\usepackage{bm}
\usepackage{float}
\usepackage{tabularx}
\usepackage{makecell}
\usepackage{array,multirow}
\usepackage{tabu}
\usepackage{cleveref}
\crefformat{section}{\S#2#1#3}
\crefformat{subsection}{\S#2#1#3}
\crefformat{subsubsection}{\S#2#1#3}
\DeclareMathOperator*{\argmax}{arg\,max}

\usepackage{tikz}
\usetikzlibrary{arrows.meta}
\usetikzlibrary{backgrounds}
\usetikzlibrary{positioning, fit, mindmap, trees, calc,tikzmark,shapes}
\usetikzlibrary{shapes.arrows, fadings, automata,tikzmark,decorations.pathreplacing,patterns}
\usepackage{pgfplots}
\usepgfplotslibrary{groupplots}
\pgfplotsset{compat=1.14}

\usepackage{graphicx}

\usepackage{sistyle}
\SIthousandsep{,}

\usepackage{xcolor}

\definecolor{riptide}{RGB}{141,211,199}
\definecolor{pale_prim}{RGB}{255,255,179}
\definecolor{lavender_gray}{RGB}{190,186,218}
\definecolor{salmon}{RGB}{242,131,107}
\definecolor{seagull}{RGB}{128,177,211}
\definecolor{rajah}{RGB}{253,180,98}
\definecolor{yellow_green}{RGB}{198,222,119}
\definecolor{classic_rose}{RGB}{252,205,229}
\definecolor{feijoa}{RGB}{178,223,138}

\definecolor{cruise}{RGB}{179,226,205}
\definecolor{apricot}{RGB}{253,205,172}
\definecolor{periwinkle}{RGB}{203,213,232}
\definecolor{snow_flurry}{RGB}{230,245,201}
\definecolor{buttermilk}{RGB}{255,242,174}

\definecolor{sundown}{RGB}{249, 180, 181}
\definecolor{spindle}{RGB}{179,205,227}
\definecolor{tea_green}{RGB}{204,235,197}
\definecolor{languid_lavender}{RGB}{222,203,228}
\definecolor{champagne}{RGB}{254,217,166}
\definecolor{cream}{RGB}{255,255,204}

\definecolor{monte_carlo}{RGB}{135,204,194}
\definecolor{melon}{RGB}{254,191,181}
\definecolor{granny_smith_apple}{RGB}{150,214,150}
\definecolor{watusi}{RGB}{254,221,207}
\definecolor{see_green}{RGB}{161,228,195}

\definecolor{moss_green}{RGB}{170,216,176}
\definecolor{opal}{RGB}{164,207,190}

\definecolor{pale_turquoise}{RGB}{172,240,242}
\definecolor{Madang}{RGB}{190,235,159}
\definecolor{pixie_green}{RGB}{183,214,170}
\definecolor{coral_andy}{RGB}{243,204,205}
\definecolor{manhattan}{RGB}{226,180,125}
\definecolor{quartz}{RGB}{219,223,238}
\definecolor{spring_sun}{RGB}{242,243,195}
\definecolor{dairy_cream}{RGB}{254,226,189}
\definecolor{surf_crest}{RGB}{205,230,208}
\definecolor{french_pass}{RGB}{195,232,246}
\definecolor{cosmos}{RGB}{248,209,210}
\definecolor{portafino}{RGB}{245,237,160}
\definecolor{sail}{RGB}{163,205,235}
\definecolor{hint_green}{RGB}{226,246,209}

\definecolor{jet_stream}{RGB}{188, 214, 210}

\definecolor{azalea}{RGB}{251, 196, 196}
\definecolor{wewak}{RGB}{244, 143, 150}
\definecolor{bittersweet}{RGB}{255,111,105}
\definecolor{sunset_orange}{RGB}{242,89,75}
\definecolor{light_coral}{RGB}{244, 127, 123}
\definecolor{carnation}{RGB}{245, 80, 86}
\definecolor{flamingo}{RGB}{237, 88, 85}
\definecolor{carmine_pink}{RGB}{231, 76, 60}
\definecolor{deep_carmine_pink}{RGB}{236, 50, 67}
\definecolor{fire_engine_red}{RGB}{210,44,41}
\definecolor{amaranth}{RGB}{234,46,73}
\definecolor{ku_crimson}{RGB}{243, 0, 25}
\definecolor{fire_engine_red}{RGB}{206, 37, 51}
\definecolor{copper_rust}{RGB}{155, 64, 74}

\definecolor{chilean_fire}{RGB}{215, 87, 44}

\definecolor{japanese_laurel}{RGB}{53, 116, 40}

\definecolor{turmeric}{RGB}{211, 178, 76}
\definecolor{saffron}{RGB}{249,193,62}
\definecolor{my_sin}{RGB}{255, 176, 59}
\definecolor{tree_poppy}{RGB}{246, 154, 27}
\definecolor{jaffa}{RGB}{240, 131, 58}
\definecolor{crusta}{RGB}{254, 127, 44}
\definecolor{tahiti_gold}{RGB}{223, 102, 36}
\definecolor{outrageous_orange}{RGB}{255, 100, 45}
\definecolor{safety_orange}{RGB}{254, 106, 0}

\definecolor{turquoise}{RGB}{41,217,194}
\definecolor{puerto_rico}{RGB}{94, 194, 166}
\definecolor{mountain_meadow}{RGB}{0, 163, 136}
\definecolor{free_speech_aquamarine}{RGB}{0, 156, 114}
\definecolor{java}{RGB}{2,190,196}

\definecolor{matisse}{RGB}{25, 104, 167}
\definecolor{shakespeare}{RGB}{85, 154, 193}
\definecolor{mona_lisa}{RGB}{246,152,134}

\definecolor{bgc}{RGB}{245,245,245}
\definecolor{tuatara}{RGB}{67, 67, 67}
\definecolor{aluminum}{RGB}{153,153,153}
\definecolor{silver}{RGB}{191,191,191}
\definecolor{platinum}{RGB}{228,228,228}
\definecolor{mercury}{RGB}{230,230,230}
\definecolor{gallery}{RGB}{240,240,240}
\definecolor{athens_gray}{RGB}{236, 240, 241}
\definecolor{ship_gray}{RGB}{77,77,77}

\definecolor{early_dawn}{RGB}{252,243,218}
\definecolor{egg_shell}{RGB}{238, 234, 215}
\definecolor{midnight}{RGB}{0, 29, 50}
\definecolor{sundown}{RGB}{249, 180, 181}
\definecolor{sun_shade}{RGB}{255, 144, 68}
\definecolor{sushi}{RGB}{117, 168, 47}
\definecolor{tomato}{RGB}{255, 97, 56}
\definecolor{ice_cold}{RGB}{169,232,220}

\definecolor{jelly_bean}{RGB}{45, 126, 150}
\definecolor{celestial_blue}{RGB}{52, 152, 219}
\definecolor{curious_blue}{RGB}{41, 128, 185}
\definecolor{french_blue}{RGB}{0, 112, 182}
\definecolor{matisse}{RGB}{25, 104, 167}

\definecolor{biscay}{RGB}{44, 62, 80}

\definecolor{cosmic_latte}{RGB}{222, 247, 229}
\definecolor{chinook}{RGB}{163, 232, 178}
\definecolor{padua}{RGB}{121, 189, 143}
\definecolor{ocean_green}{RGB}{79, 176, 112}
\definecolor{pastel_green}{RGB}{107, 227, 135}
\definecolor{chateau_green}{RGB}{69, 191, 85}
\definecolor{RoyalBlue}{RGB}{69, 191, 85}
\definecolor{pigment_green}{RGB}{0, 175, 79}
\definecolor{fern}{RGB}{101,197,117}
\definecolor{killarney}{RGB}{56, 113, 66}
\definecolor{viridian}{RGB}{70, 137, 102}

\usepackage{microtype}
\definecolor{amaranth}{rgb}{0.9, 0.17, 0.31}
\definecolor{kellygreen}{rgb}{77, 186, 23}
\definecolor{azure}{rgb}{0.0, 0.5, 1.0}
\definecolor{gred}{rgb}{0.9, 0.17, 0.31}
\definecolor{gblue}{rgb}{0.0, 0.5, 1.0}
\definecolor{gyellow}{RGB}{244,180,0}
\definecolor{ggreen}{rgb}{0.3, 0.73, 0.09}
\definecolor{ggrey}{RGB}{115,115,115}

\newcommand{\error}[1]{\textcolor{gred}{\textbf{#1}}} 

\aclfinalcopy 

\title{Diversifying Dialogue Generation with Non-Conversational Text}

\author{Hui Su$^{1}$\thanks{\hspace{1.5 mm}Equal contribution.}, 
Xiaoyu Shen$^2\footnotemark[1]\hspace{1.5 mm}$\\
\textbf{Sanqiang Zhao$^3$, Xiao Zhou$^1$, Pengwei Hu$^{4}$, Randy Zhong$^{1}$, Cheng Niu$^{1}$ and Jie Zhou$^{1}$}\\
$^1$Pattern Recognition Center, Wechat AI, Tencent Inc, China\\
$^2$MPI Informatics \& Spoken Language Systems (LSV), Saarland Informatics Campus\\
$^3$University of Pittsburgh  \hspace{5 mm} $^4$The Hong Kong Polytechnic University, Hong Kong\\
\tt{aaronsu@tencent.com,xshen@mpi-inf.mpg.de}}

\date{}

\begin{document}
\maketitle
\begin{abstract}
Neural network-based sequence-to-sequence (seq2seq) models strongly suffer from the low-diversity problem when it comes to open-domain dialogue generation. As bland and generic utterances usually dominate the frequency distribution in our daily chitchat, avoiding them to generate more interesting responses requires complex data filtering, sampling techniques or modifying the training objective. In this paper, we propose a new perspective to diversify dialogue generation by leveraging \emph{non-conversational} text. Compared with bilateral conversations, non-conversational text are easier to obtain, more diverse and cover a much broader range of topics. We collect a large-scale non-conversational corpus from multi sources including forum comments, idioms and book snippets. We further present a training paradigm to effectively incorporate these text via iterative back translation. The resulting model is tested on two conversational datasets and is shown to produce significantly more diverse responses without sacrificing the relevance with context. 
\end{abstract}

\section{Introduction}

Seq2seq models have achieved impressive success in a wide range of text generation tasks. In open-domain chitchat, however, people have found the model tends to strongly favor short, generic responses like ``I don't know" or ``OK"~\cite{vinyals2015neural,shen2017estimation}. The reason lies in the extreme one-to-many mapping relation between every context and its potential responses~\cite{zhao2017learning,su2018dialogue}. Generic utterances, which can be in theory paired with most context, usually dominate the frequency distribution in the dialogue training corpus and thereby pushes the model to blindly produce these safe, dull responses ~\cite{su2019improving,csaky-etal-2019-improving}

\begin{table}[!t]
    	 \small
    	\centering	
    		\scalebox{0.9}
    		{
    	\begin{tabular}{l|l}
    	  \multicolumn{2}{c}{\textbf{Conversational Text}}\\\hline 
    	\textbf{Context}& \begin{CJK*}{UTF8}{gbsn}暗恋的人却不喜欢我\end{CJK*} \\(Translation)& The one I have a crush on doesn't like me. \\ \hline
    	\multirow{2}{1.5cm}{\textbf{Response}} & \begin{CJK*}{UTF8}{gbsn} 摸摸头\end{CJK*} \\ & Head pat.\\ \hline

    	\multicolumn{2}{c}{\rule{0pt}{10pt}\textbf{Non-Conversational Text}}\\\hline
    	
    	\multirow{3}{1.5cm}{\textbf{Forum Comments}}& \begin{CJK*}{UTF8}{gbsn} 暗恋这碗酒，谁喝都会醉啊 \end{CJK*} \\& Crush is an alcoholic drink, whoever drinks\\& it will get intoxicated.\\ \hline
    	\multirow{2}{1.5cm}{\textbf{Idiom}} &  \begin{CJK*}{UTF8}{gbsn} 何必等待一个没有结果的等待\end{CJK*} \\& Why wait for a result without hope \\ \hline
    	\multirow{3}{1.5cm}{\textbf{Book Snippet}} &\begin{CJK*}{UTF8}{gbsn} 真诚的爱情之路永不会是平坦的\end{CJK*} \\& The course of true love never did run smooth \\ & (From \emph{A Midsummer Night's Dream})\\ \hline
    	\end{tabular}
}
    	\caption{\small A daily dialogue and non-conversational text from three sources. The contents of non-conversational text can be potentially utilized to enrich the response generation.}	\vspace{-5mm}
    	\label{tab:dialog}
    \end{table}

Current solutions can be roughly categorized into two classes: (1) Modify the seq2seq itself to bias toward diverse responses~\cite{li2015diversity,shen2019select}. However, the model is still trained on the \emph{limited dialogue corpus} which restricts its power at covering broad topics in open-domain chitchat. (2) Augment the training corpus with extra information like structured world knowledge, personality or emotions~\cite{li2016persona,dinan2018wizard}, which requires \emph{costly human annotation}.

In this work, we argue that training only based on conversational corpus can greatly constrain the usability of an open-domain chatbot system since many topics are not easily available in the dialogue format. With this in mind, we explore a cheap way to diversify dialogue generation by utilizing large amounts of \emph{non-conversational text}. Compared with bilateral conversations, non-conversational text covers a much broader range of topics, and can be easily obtained without further human annotation from multiple sources like forum comments, idioms and book snippets. More importantly, non-conversational text are usually \emph{more interesting and contentful} as they are written to convey some specific personal opinions or introduce a new topic, unlike in daily conversations where people often \emph{passively} reply to the last utterance. As can be seen in Table~\ref{tab:dialog}, the response from the daily conversation is a simple comfort of ``Head pat". Non-conversational text, on the contrary, exhibit diverse styles ranging from casual wording to poetic statements, which we believe can be potentially utilized to enrich the response generation.

To do so, we collect a large-scale corpus containing over 1M non-conversational utterances from multiple sources. To effectively integrate these utterances, we borrow the back translation idea from unsupervised neural machine translation~\cite{sennrich2016improving,lample2018phrase} and treat the collected utterances as unpaired responses. We first pre-train the forward and backward transduction model on the parallel conversational corpus. The forward and backward model are then iteratively tuned to find the optimal mapping relation between conversational context and non-conversational utterances~\cite{cotterell2018explaining}. By this means, the content of non-conversational utterances is gradually distilled into the dialogue generation model~\cite{kim2016sequence}, enlarging the space of generated responses to cover not only the original dialogue corpus, but also the wide topics reflected in the non-conversational utterances.

We test our model on two popular Chinese conversational datasets weibo~\cite{shang2015neural} and douban~\cite{wu2017sequential}. We compare our model against retrieval-based systems, style-transfer methods and several seq2seq variants which also target the diversity of dialogue generation. Automatic and human evaluation show that our model significantly improves the responses' diversity both semantically and syntactically without sacrificing the relevance with context, and is considered as most favorable judged by human evaluators~\footnote{Code and dataset available at \url{https://github.com/chin-gyou/Div-Non-Conv}}.
\section{Related Work}
The tendency to produce generic responses has been a long-standing problem in seq2seq-based open-domain dialogue generation~\cite{vinyals2015neural,li2015diversity}. Previous approaches to alleviate this issue can be grouped into two classes. 

The first class resorts to modifying the seq2seq architecture itself. For example, \citet{shen2018nexus,zhang2018generating} changes the training objective to mutual information maximization and rely on continuous approximations or policy gradient to circumvent the non-differentiable issue for text. \citet{li2016deep,serban2017deep} treat open-domain chitchat as a reinforcement learning problem and manually define some rewards to encourage long-term conversations. There is also research that utilizes latent variable sampling~\cite{serban2016hierarchical,shen2018improving,shen2019improving}, adversarial learning~\cite{li2017adversarial,su2018dialogue}, replaces the beam search decoding with a more diverse sampling strategy~\cite{li2016simple,holtzman2019curious} or applies reranking to filter generic responses~\cite{li2015diversity,wang2017steering}. All of the above are still trained on the original dialogue corpus and thereby cannot generate out-of-scope topics.

The second class seeks to bring in extra information into existing corpus like structured knowledge~\cite{zhao2018comprehensive,ghazvininejad2018knowledge,dinan2018wizard}, personal information~\cite{li2016persona,zhang2018personalizing} or emotions~\cite{shen2017conditional,zhou2018emotional}. However, corpus with such annotations can be extremely costly to obtain and is usually limited to a specific domain with small data size. Some recent research started to do dialogue style transfer based on personal speeches or TV scripts~\cite{niu2018polite,gao2019structuring,su2019personalized}. Our motivation differs from them in that we aim at enriching general dialogue generation with abundant non-conversational text instead of being constrained on one specific type of style.

Back translation is widely used in unsupervised machine translation~\cite{sennrich2016improving,lample2017unsupervised,artetxe2017unsupervised} and has been recently extended to similar areas like style transfer~\cite{subramanian2018multiple}, summarization~\cite{zhao2019unsupervised} and data-to-text~\cite{chang2020unsupervised}. To the best of our knowledge, it has never been applied to dialogue generation yet. Our work treats the context and non-conversational text as unpaired source-target data. The back-translation idea is naturally adopted to learn the mapping between them. The contents of non-conversational text can then be effectively utilized to enrich the dialogue generation.
\section{Dataset}
\begin{table}
    \centering
    \begin{adjustbox}{max width=\linewidth}
    \begin{tabular}{l c c }
    \toprule
    Resources & Size & Avg. length \\
    \midrule
    Comments & 781,847& 21.0 \\
    Idioms & 51,948 &  18.7  \\
    Book Snippets & 206,340 & 26.9  \\
    \bottomrule
    \end{tabular}
    \end{adjustbox}
 \caption{\label{tab:mono_data}\small Statistics of Non-Conversational Text.}
\end{table}
We would like to collect non-conversational utterances that stay close with daily-life topics and can be potentially used to augment the response space. The utterance should be neither too long nor too short, similar with our daily chitchats. Therefore, we collect data from the following three sources:
\begin{enumerate}
    \item Forum comments. We collect comments from zhihu~\footnote{\url{https://www.zhihu.com}}, a popular Chinese forums. Selected comments are restricted to have more than 10 likes and less than 30 words~\footnote{The posts are usually very long, describing a specific social phenomenon or news event, so building parallel conversational corpus from post-comment pairs is difficult. Nonetheless, these high-liked comments are normally high-quality themselves and can be used to augment the response space.}.
    \item Idioms. We crawl idioms, famous quotes, proverbs and locutions from several websites. These phrases are normally highly-refined and graceful, which we believe might provide a useful augmentation for responses.
    \item Book Snippets. We select top 1,000 favorite novels or prose from wechat read~\footnote{\url{https://weread.qq.com/}}. Snippets highlighted by readers, which are usually quintessential passages, and with the word length range 10-30 are kept.
\end{enumerate}
We further filter out sentences with offensive or discriminative languages by phrase matching against a large blocklist. The resulting corpus contains over 1M utterances. The statistics from each source are listed in Table~\ref{tab:mono_data}.
\section{Approach}
\label{sec: approach}
\begin{figure*}[!ht]
\centering
\centerline{\includegraphics[width=15cm]{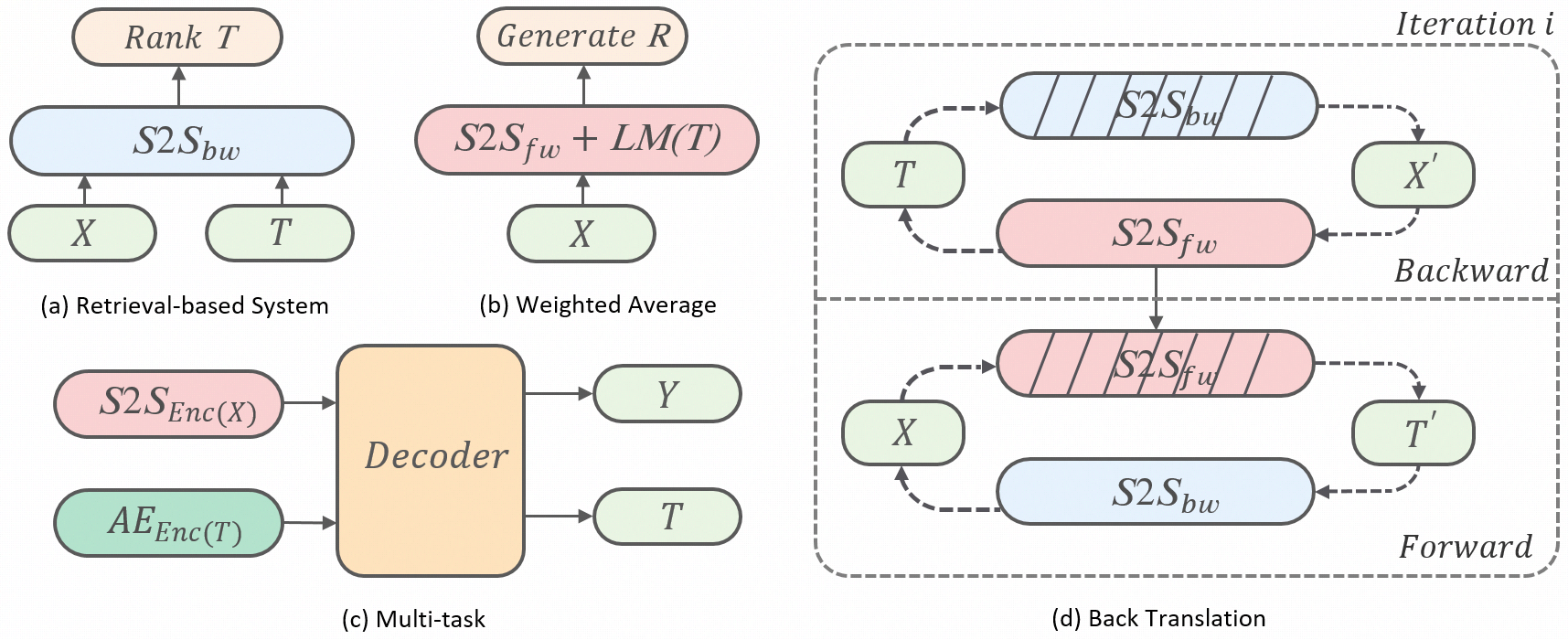}}
\caption{\small Comparison of four approaches leveraging the non-conversational text. $S2S_{fw}$, $S2S_{bw}$ and $LM$ indicate the forward, backward seq2seq and language model respectively. (d) visualizes the process of one iteration for the back translation approach. Striped component are not updated in each iteration.}
\label{fig:model}
\end{figure*}
Let $\mathcal{D}=\{(X_1, Y_1),(X_2, Y_2),\ldots,(X_N,Y_N)\}$ denote the parallel conversational corpus. $X_i$ is the context and $Y_i$ is the corresponding response. $\mathcal{D}_T=\{T_1, T_2,\ldots,T_M\}$ denotes our collected corpus where $T_i$ is a non-conversational utterance. As the standard seq2seq model trained only on $\mathcal{D}$ tends to generate over-generic responses, our purpose is to diversify the generated responses by leveraging the non-conversational corpus $\mathcal{D}_T$, which are semantically and syntactically much richer than responses contained in $\mathcal{D}$. In the following section, we first go through several baseline systems, then introduce our proposed method based on back translation.
\subsection{Retrieval-based System}
\label{sec: ret}
The first approach we consider is a retrieval-based system that considers all sentences contained in $\mathcal{D}_T$ as candidate responses. As the proportion of generic utterances in $\mathcal{D}_T$ is much lower than that in $\mathcal{D}$, the diversity will be largely improved. Standard retrieval algorithms based on context-matching~\cite{wu2017sequential,bartl2017retrieval} fail to apply here since non-conversational text does not come with its corresponding context. Therefore, we train a backward seq2seq model on the parallel conversational corpus $\mathcal{D}$ to maximize $p(X_i|Y_i)$. The score assigned by the backward model, which can be seen as an estimation of the point-wise mutual information, is used to rank the responses~\cite{li2015diversity}~\footnote{The backward seq2seq model measures the context relevance better than forward models since the latter highly biases generic utterances~\cite{li2015diversity,zhang2018generating}}.

The major limitation of the retrieval-based system is that it can only produce responses from a finite set of candidates. The model can work well only if an appropriate response already exists in the candidate bank. Nonetheless, due to the large size of the non-conversational corpus, this approach is a very strong baseline.
\subsection{Weighted Average}
\label{sec: weight}
The second approach is to take a weighted average score of a seq2seq model trained on $\mathcal{D}$ and a language model trained on $\mathbf{D}_T$ when decoding responses. The idea has been widely utilized on domain adaptation for text generation tasks~\cite{koehn2007experiments,wang2017steering,niu2018polite}. In our scenario, basically we hope the generated responses could share the diverse topics and styles of the non-conversational text, yet stay relevant with the dialogue context. The seq2seq model $S2S$ is trained on $\mathcal{D}$ as an indicator of how relevant each response is with the context. A language model $\mathcal{L}$ is trained on $\mathcal{D}_T$ to measure how the response matches the domain of $\mathcal{D}_T$. The decoding probability for generating word $w$ at time step $t$ is assigned by:
\begin{equation}
\label{eq: weight}
    p_t(w) = \alpha S2S_t(w) + (1-\alpha) L_t(w)
\end{equation}
where $\alpha$ is a hyperparameter to adjust the balance between the two. Setting $\alpha=1$ will make it degenerate into the standard seq2seq model while $\alpha=0$ will totally ignore the dialoge context.
\subsection{Multi-task}
\label{sec: multi}
The third approach is based on multi-task learning. A seq2seq model is trained on the parallel conversational corpus $\mathcal{D}$ while an autoencoder model is trained on the non-parallel monologue data $\mathcal{D}_T$. Both models share the decoder parameters to facilitate each other. The idea was first experimented on machine translation in order to leverage large amounts of target-side monolingual text~\cite{luong2015multi,sennrich2016improving}. \citet{luan2017multi} extended it to conversational models for speaker-role adaptation. The intuition is that by tying the decoder parameters, the seq2seq and autoencoder model can learn a shared latent space between the dialogue corpus and non-conversational text. When decoding, the model can generate responses with features from both sides.

\subsection{Back Translation}
\label{sec: bt}
Finally, we consider the back translation technique commonly used for unsupervised machine translation~\cite{artetxe2017unsupervised,lample2017unsupervised}. The basic idea is to first \emph{initialize} the model properly to provide a good starting point, then iteratively perform \emph{backward} and \emph{forward} translation to learn the correspondence between context and unpaired non-conversational utterances.
\paragraph{Initialization}
Unlike unsupervised machine translation, the source and target side in our case come from the same language, and we already have a parallel conversational corpus $\mathcal{D}$, so we can get rid of the careful embedding alignment and autoencoding steps as in \citet{lample2018phrase}. For the initialization, we simply train a forward and backward seq2seq model on $\mathcal{D}$. The loss function is:
\begin{equation}
\label{eq: init}
    \mathbb{E}_{X_i, Y_i \sim \mathcal{D}} -\log P_{f}(Y_i|X_i) - \log P_{b}(X_i|Y_i)
\end{equation}
where $P_f$ and $P_b$ are the decoding likelihood defined by the forward and backward seq2seq model respectively. We optimize Eq.~\ref{eq: init} until convergence. Afterwards, the forward and backward seq2seq can learn the backbone mapping relation between a context and its response in a conversational structure.
\paragraph{Backward}
After the initialization, we use the backward seq2seq to create pseudo parallel training examples from the non-conversational text $\mathcal{D}_T$. The forward seq2seq is then trained on the pseudo pairs. The objective is to minimize:
\begin{equation}
\label{eq: backward}
\begin{split}
    \mathbb{E}_{T_i \sim \mathcal{D}_T}& -\log P_{f}(T_i|b(T_i))\\
    b(T_i)&= \argmax_{u} P_b(u|T_i)
\end{split}
\end{equation}
where we approximate the $\argmax$ function by using a beam search decoder to decode from the backward model $P_b(u|T_i)$. Because of the non-differentiability of the $\argmax$ operator, the gradient is only passed through $P_f$ but not $P_b$~\footnote{As also noted in \citet{lample2018phrase}, backpropagating further through $P_b$ brings no improvement.}.

As $P_b$ is already well initialized by training on the parallel corpus $\mathcal{D}$, the back-translated pseudo pair $\{b(T_i), T_i\}$ can roughly follow the typical human conversational patterns. Training $P_f$ on top of them will encourage the forward decoder to generate utterances in the domain of $T_i$ while maintaining coherent as a conversation.
\paragraph{Forward}
The forward translation follows a similar step as back translation. The forward seq2seq $P_f$ translates context into a response, which in return form a pseudo pair to train the backward model $P_b$. The objective is to minimize:
\begin{equation}
\label{eq: forward}
\begin{split}
    \mathbb{E}_{X_i \sim \mathcal{D}}& -\log P_{b}(X_i|f(X_i))\\
    f(X_i)&= \argmax_{v} P_f(v|X_i)
\end{split}
\end{equation}
where the $\argmax$ function is again approximated with a beam search decoder and the gradient is only backpropagated through $P_b$. Though $X_i$ has its corresponding $Y_i$ in $\mathcal{D}$, we drop $Y_i$ and instead train on forward translated pseudo pairs $\{X_i,f(X_i)\}$. As $P_f$ is trained by leveraging data from $\mathcal{D}_T$, $f(X_i)$ can have superior diversity compared with $Y_i$.

The encoder parameters are shared between the forward and backward models while decoders are separate. The backward and forward translation are iteratively performed to close the gap between $P_f$ and $P_b$~\cite{hoang2018iterative,cotterell2018explaining}. The effects of non-conversational text are strengthened after each iteration. Eventually, the forward model will be able to produce diverse responses covering the wide topics in $\mathcal{D}_T$. Algorithm~\ref{algorithm} depicts the training process.

\begin{algorithm}[t]
    (\textbf{Inilialization})
    Train by minimizing Eq.~\ref{eq: init} until convergence\;
 \For{i=1 to N}{
  (\textbf{Backward}) Train by minimizing Eq.~\ref{eq: backward} until convergence\;
   (\textbf{Forward}) Train by minimizing Eq.~\ref{eq: forward} until convergence\;
 }
 \caption{\label{algorithm}\small Model Training Process}
\end{algorithm}

\section{Experiments}
\subsection{Datasets}
We conduct our experiments on two Chinese dialogue corpus Weibo~\cite{shang-etal-2015-neural} and Douban~\cite{wu2017sequential}. Weibo~\footnote{\url{http://www.weibo.com/}} is a popular Twitter-like microblogging service in China, on which a user can post short messages, and other users make comment on a published post. The post-comment pairs are crawled as short-text conversations. Each utterance has 15.4 words on average and the data is split into train/valid/test subsets with 4M/40k/10k utterance pairs. Douban~\footnote{\url{https://www.douban.com/group}} is a Chinese social network service where people can chat about different topics online. The original data contains 1.1M multi-turn conversations. We split them into two-turn context-response pairs, resulting in 10M train, 500k valid and 100K test samples.
\subsection{General Setup}
For all models, we use a two-layer LSTM~\cite{hochreiter1997long} encoder/decoder structure with hidden size 500 and word embedding size 300. Models are trained with Adam optimizer~\cite{kingma2014adam} with an initial learning rate of 0.15. We set the batch size as 256 and use the gradients clipping of 5. We build out vocabulary with character-based segmentation for Chinese. For non-Chinese tokens, we simply split by space and keep all unique tokens that appear at least 5 times. Utterances are cut down to at most 50 tokens and fed to every batch. We implement our models based on the OpenNMT toolkit~\cite{klein2017opennmt} and other hyperparameters are set as the default values.
\subsection{Compared Models}
We compare our model with the standard seq2seq and four popular variants which were proposed to improve the diversity of generated utterances. All of them are trained only on the parallel conversational corpus:
\paragraph{Standard} The standard seq2seq with beam search decoding (size 5).
\paragraph{MMI} The maximum mutual information decoding which reranks the decoded responses with a backward seq2seq model~\cite{li2015diversity}. The hyperparameter $\lambda$ is set to 0.5 as suggested. 200 candidates
per context are sampled for re-ranking
\paragraph{Diverse Sampling} The diverse beam search strategy proposed in \citet{vijayakumar2016diverse} which explicitly controls for the exploration and
exploitation of the search space. We set the number of groups as 5, $\lambda=0.3$ and use the Hamming diversity as the penalty function as in the paper.
\paragraph{Nucleus Sampling} Proposed in \citet{holtzman2019curious}, it allows for diverse sequence generations. Instead of decoding with a fixed beam size, it samples text from the dynamic nucleus. We use the default configuration and set $p=0.9$.
\paragraph{CVAE} The conditional variational autoencoder~\cite{serban2016hierarchical,zhao2017learning} which injects diversity by imposing stochastical latent variables. We use a latent variable with dimension 100 and utilize the KL-annealing strategy with step 350k and a word drop-out rate of 0.3 to alleviate the posterior collapse problem~\cite{bowman2016generating}.

Furthermore, we compare the 4 approaches mentioned in \cref{sec: approach} which incorporate the collected non-conversational text:
\paragraph{Retrieval-based}(\cref{sec: ret}) Due to the large size of the non-conversational corpus, exact ranking is extremely slow. Therefore, we first retrieve top 200 matched text with elastic search based on the similarity of Bert embeddings~\cite{devlin2019bert}. Specifically, we pass sentences through Bert and derive a fixed-sized vector by averaging the outputs from the second-to-last layer~\cite{may2019measuring}~\footnote{\url{https://github.com/hanxiao/bert-as-service}}. The 200 candidates are then ranked with the backward score~\footnote{This makes it similar to MMI reranking, whose 200 candidates are from seq2seq decodings instead of top-matched non-conversational utterances.}.
\paragraph{Weighted Average}(\cref{sec: weight}) We set $\lambda=0.5$ in eq.~\ref{eq: weight}, which considers context relevance and diversity with equal weights.
\paragraph{Multi-task}((\cref{sec: multi})) We concatenate each context-response pair with a non-conversational utterance and train with a mixed objective of seq2seq and autoencoding (by sharing the decoder).
\paragraph{Back Translation}(\cref{sec: bt}) We perform the iterative backward and forward translation 4 times for both datasets. We observe the forward cross entropy loss converges after 4 iterations.

\section{Results}
As for the experiment results, we report the automatic and human evaluation in \cref{sec: auto} and \cref{sec: human} respectively. Detailed analysis are shown in \cref{sec: analysis} to elaborate the differences among model performances and some case studies.

\subsection{Automatic Evaluation}
\label{sec: auto}
Evaluating dialogue generation is extremely difficult. Metrics which measure the word-level overlap like BLEU~\cite{papineni2002bleu} have been widely used for dialogue evaluation. However, these metrics do not fit into our setting well as we would like to diversify the response generation with an external corpus, the generations will inevitably differ greatly from the ground-truth references in the original conversational corpus. Though we report the BLEU score anyway and list all the results in Table~\ref{tab: auto}, it is worth mentioning that the BLEU score itself is by no means a reliable metric to measure the quality of dialogue generations.
\begin{table*}[t!]
  \small
  \begin{center}
    \begin{tabularx}{0.95\linewidth}{l|ccccc|ccccc}
    \hline 
     \textbf{Metrics} & 
     \multicolumn{5}{c|}{\textbf{Weibo}} & \multicolumn{5}{c}{\textbf{Douban}}  \\ 
     \textbf{Model}& \textbf{BLEU-2} &  \textbf{Dist-1} & \textbf{Dist-2} &   \textbf{Ent-4} & \textbf{Adver}&\textbf{BLEU-2} &  \textbf{Dist-1} & \textbf{Dist-2}  & \textbf{Ent-4}  & \textbf{Adver} \\
     \hline
     \textbf{\textsc{Standard}} & 0.0165  & 0.018 & 0.050  & 5.04 & 0.30 & 0.0285 & 0.071 & 0.206  & 7.55 & 0.19 \\
     \textbf{\textsc{MMI}} & 0.0161 & 0.025 & 0.069   & 5.98  & 0.42 & 0.0263 & 0.143 & 0.363  & 7.60  & \textbf{0.31}  \\
     \textbf{\textsc{Diverse}}  & 0.0175 & 0.019  & 0.054   & 6.20  & 0.38 & 0.0298 & 0.130 & 0.358  & 7.51   & 0.25  \\
     \textbf{\textsc{Nucleus}} & \textbf{0.0183} & \textbf{0.027}  & \textbf{0.074} & \textbf{7.41}  & \textbf{0.43} & \textbf{0.0312}  & 0.141 &  0.402  & \textbf{7.93} & 0.30  \\
     \textbf{\textsc{CVAE}} & 0.0171 & 0.023  & 0.061  & 6.63  & 0.36  & 0.0287 & \textbf{0.169} &  \textbf{0.496}  & 7.80  & 0.29 \\
    \hline
    \textbf{\textsc{Retrieval}}  & 0.0142 & \textbf{0.198} & \textbf{0.492}   & \textbf{12.5}  & 0.13 &\textbf{ 0.0276} & \textbf{0.203} & \textbf{0.510} & \textbf{13.3}  & \textbf{0.17} \\
     \textbf{\textsc{Weighted}} &\textbf{ 0.0152}  & 0.091  & 0.316   & 9.26  & 0.22 & 0.0188 & 0.172 & 0.407  & 8.73   & 0.14 \\
     \textbf{\textsc{Multi}}  & 0.0142 & 0.128  & 0.348  & 8.98  & \textbf{0.27}  & 0.0110 & 0.190 & 0.389  & 8.26& 0.16 \\    \hline
      \textbf{\textsc{BT (Iter=1)}}  & \textbf{0.0180} & 0.046  & 0.171  & 7.64  & 0.19  & \textbf{0.0274} &  0.106 & 0.313  & 8.16   & 0.15 \\
     \textbf{\textsc{BT (Iter=4)}} & 0.0176  & \textbf{0.175}   & \textbf{0.487}  & \textbf{11.2} & \textbf{0.35}  & 0.0269 &  \textbf{0.207} & \textbf{0.502}  & \textbf{11.0}   & \textbf{0.25} \\
     \hline
      \textbf{\textsc{Human}} & - & 0.171    & 0.452 & 9.23  & 0.88  & - & 0.209 & 0.514  &11.3   & 0.85 \\\hline
    \end{tabularx}
  \end{center}
   \caption{\small Automatic evaluation on Weibo and Douban datasets. Upper areas are models trained only on the conversational corpus. Middle areas are baseline models incorporating the non-conversational corpus. Bottom areas are our model with different number of iterations. Best results in every area are \textbf{bolded}.}
     \label{tab: auto}%
     \vspace{-0.3cm}
\end{table*}%
\paragraph{Diversity}
Diversity is a major concern for dialogue generation.  Same as in \cite{li2015diversity}, we measure the diversity by the ratio of distinct unigrams (\textbf{Dist-1}) and bigrams (\textbf{Dist-2}) in all generated responses. As the ratio itself ignores the frequency distribution of n-grams, we further calculate the entropy value for the empirical distribution of n-grams~\cite{zhang2018generating}. A larger entropy indicates more diverse distributions. We report the entropy of four-grams (\textbf{Ent-4}) in Table~\ref{tab: auto}. Among models trained only on the conversational corpus, the standard seq2seq performed worst as expected. All different variants improved the diversity more or less. Nucleus sampling and CVAE generated most diverse responses, especially Nucleus who wins on 6 out of the 8 metrics. By incorporating the non-conversational corpus, the diversity of generated responses improves dramatically. The retrieval-based system and our model perform best, in most cases even better than human references. This can happen as we enrich the response generation with external resources. The diversity would be more than the original conversational corpus. Weighted-average and multi-task models are relatively worse, though still greatly outperforming models trained only on the conversational corpus. We can also observe that our model improves over standard seq2seq only a bit after one iteration. As more iterations are added, the diversity improves gradually.
\paragraph{Relevance}
Measuring the context-response relevance automatically is tricky in our case. The typical way of using scores from forward or backward models as in \citet{li2016neural} is not suitable as our model borrowed information from extra resources. The generated responses are out-of-scope for the seq2seq model trained on only on the conversational corpus and thus would be assigned very low scores. Apart from the BLEU-2 score, we further evaluate the relevance by leveraging an adversarial discriminator~\cite{li2017adversarial}. As has been shown in previous research, discriminative models are generally less biased to high-frequent utterances and more robust against their generative counterparts~\cite{lu2017best,luo2018discriminability}. The discriminator is trained on the parallel conversational corpus distinguish correct responses from randomly sampled ones. We encode the context and response separately with two different LSTM neural networks and output a binary signal indicating relevant or not~\footnote{In our experiment, the discriminator performs reasonably well in the 4 scenarios outlined in \citet{li2017adversarial} and thus can be considered as a fair evaluation metric.}. The relevance score is defined as the success rate that the model fools the adversarial classifier into believing its generations (\textbf{Adver} in Table~\ref{tab: auto}). The retrieval-based model, who generates the most diverse generations, achieve the lowest score as for relevance with context. The restriction that it can only select from a set of fixed utterances do affect the relevance a lot~\footnote{The fact that we only rank on 200 most similar utterances might also affect. We tried increasing the size to 1,000 but observe no tangible improvement. The candidate size required for a decent relevance score can be unbearably large.}. Note that \emph{the discriminator is also trained on the same bilateral conversational corpus, putting our model into a naturally disadvantageous place due to the incorporation of out-of-scope non-conversational text.} Nonetheless, our model still achieves competitive relevance score even compared with models trained only on the conversational corpus. This suggests our model does learn the proper patterns in human conversations instead of randomly synthesizing diverse generations.
  \begin{table}[!hbtp] \addtolength{\tabcolsep}{-2pt}  
      \footnotesize
      \centering
      \begin{tabularx}{0.95\linewidth}{l|ccc|ccc}
          \hline
             \textbf{Metrics} & 
     \multicolumn{3}{c|}{\textbf{Weibo}} & \multicolumn{3}{c}{\textbf{Douban}}  \\ 
          \textbf{Model} & \textbf{Rel} &\textbf{Inter}& \textbf{Flu} & \textbf{Rel} &\textbf{Inter}& \textbf{Flu} \\ \hline 
          STANDARD & 0.32 & 0.11 & 0.76 & 0.26 & 0.13 & 0.82 \\
          NUCLEUS & \textbf{0.46} & 0.19 & \textbf{0.78} & 0.38 & 0.21 & \textbf{0.83} \\ \hline 
          RETRIEVAL & 0.12 & 0.35 & - & 0.09 & 0.32 & - \\ 
          WEIGHTED & 0.19& 0.14& 0.52 & 0.15 & 0.17& 0.46 \\ 
          MULTI & 0.25& 0.21& 0.70 & 0.22 & 0.23& 0.66 \\ 
          BT (ITER=4) & 0.43 & \textbf{0.37} & 0.77 & \textbf{0.39} & \textbf{0.48} & 0.80 \\ \hline
      \end{tabularx}
     \caption{\label{tab:human}\small Human Evaluation Results}
  \end{table}
  \begin{table*}[t]
\centering
\begin{tabular}{l|l}
\hline
Context                  & \makecell[l]{\begin{CJK*}{UTF8}{gbsn}一 直 单 身 怎 么 办\end{CJK*}(Being always single, what should I do?)}\\
\hline
Response                    & \makecell[l]{ {\begin{CJK*}{UTF8}{gbsn}勇 敢 一 点 多 去 加 好 友 啊\end{CJK*} }(Be brave and add more people to friends.) }\\ 
\hline
\multirow{3}{*}{Generation} & \makecell[l]{{[}Iteration 0{]}: \begin{CJK*}{UTF8}{gbsn}不知道该怎么办 \end{CJK*} (I don't know what to do.) } \\ \cline{2-2} 
&\makecell[l]{{[}Iteration 1{]}: \begin{CJK*}{UTF8}{gbsn}单 身 不 可 怕 ， 单 身 不 可 怕(Being single is nothing, being single is nothing.)\end{CJK*}} \\ \cline{2-2} 
& \makecell[l]{{[}Iteration 4{]}:
\begin{CJK*}{UTF8}{gbsn}斯 人 若 彩 虹 ，遇 上 方 知 有\end{CJK*}(Every once in a while you find someone who's\\ iridescent, and when you do, nothing will ever compare.)} \\ 
\hline
\end{tabular}
\caption{\small Example of response generation in different iterations.}
\label{table:iter}
\end{table*}
\subsection{Human Evaluation}
\label{sec: human}
Apart from automatic evaluations, we also employed crowdsourced judges to evaluate the quality of generations for 500 contexts of each dataset. We focus on evaluating the generated responses regarding the (1) relevance: if they coincide with the context (\textbf{Rel}), (2) interestingness: if they are interesting for people to continue the conversation (\textbf{Inter}) and (3) fluency: whether they are fluent by grammar (\textbf{Flu})~\footnote{We do not evaluate the retrieval-based model for the fluency score as the retrieved utterances are fluent by construct.}. Each sample gets one point if judged as yes and zero otherwise. Each pair is judged by three participants and the score supported by most people is adopted. The averaged scores are summarized in Table~\ref{tab:human}. We compare the standard seq2seq model, nucleus sampling which performs best among all seq2seq variants, and the four approaches leveraging the non-conversational text. All models perform decently well as for fluency except the weighted average one. The scores for diversity and relevance generally correlate well with the automatic evaluations. Overall the back-translation model are competitive with respect to fluency and relevance, while generating much more interesting responses to human evaluators. It also significantly outperforms the other three baseline approaches in its capability to properly make use of the non-conversational corpus.

\subsection{Analysis}
\label{sec: analysis}

\pgfplotsset{
axis background/.style={fill=gallery},
grid=both,
  xtick pos=left,
  ytick pos=left,
  tick style={
    major grid style={style=white,line width=1pt},
    minor grid style=bgc,
    draw=none
    },
  minor tick num=1,
  ymajorgrids,
	major grid style={draw=white},
	y axis line style={opacity=0},
	tickwidth=0pt,
}

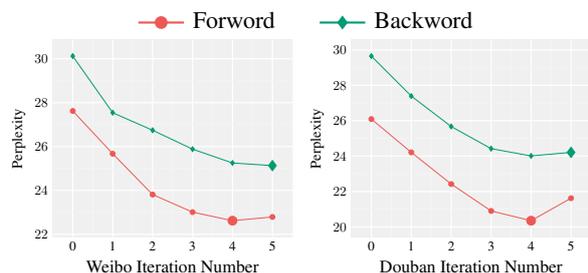
\begin{figure}[ht]
\centering
    \begin{tikzpicture}[scale=0.46]
	\begin{groupplot}[
	    group style={group size=2 by 1,
	        horizontal sep = 48pt}, 
	    xlabel=\Large  Weibo Iteration Number,
        ylabel=\large Perplexity,
        xticklabels={0,1,2,3,4,5},
        xtick={1,2,3,4,5,6},
        ymajorgrids,
        major grid style={draw=white},
        y axis line style={opacity=0},
        tickwidth=0pt,
        yticklabel style={
        /pgf/number format/fixed,
        /pgf/number format/precision=5
        },
        scaled y ticks=false,
        every axis title/.append style={at={(0.1,0.8)},font=\bfseries}
	    ]
		\nextgroupplot[
		legend style = {
		  font=\small,
          draw=none, 
          fill=none,
          column sep = 1pt, 
          /tikz/every even column/.append style={column sep=5mm},
          legend columns = -1, 
          legend to name = grouplegend},
		]

		\addplot[thick,color=flamingo,mark=*] coordinates {
          (1, 27.62)
          (2, 25.67 )
          (3, 23.81)
          (4, 23.01)
          (5, 22.62)
          (6, 22.79)
        }; \addlegendentry{\textcolor{black}{Forword}}
         \addplot[thick,color=free_speech_aquamarine,mark=diamond*] coordinates {
          (1, 30.12)
          (2, 27.54)
          (3, 26.74)
          (4, 25.88)
          (5, 25.25)
          (6, 25.13)
        }; \addlegendentry{\textcolor{black}{Backword}}

        \addplot[mark size=3.6pt, color=flamingo, mark=*,
        ] coordinates {
          (5, 22.62)
        }; 
         \addplot[mark size=4.2pt, color=free_speech_aquamarine, mark=diamond*] coordinates {
          (6, 25.13)
        };

        \nextgroupplot[
	    xlabel=\Large  Douban Iteration Number]
		\addplot[thick,color=flamingo,mark=*] coordinates {
          (1, 26.09)
          (2, 24.21)
          (3, 22.42)
          (4, 20.90)
          (5, 20.35 )
          (6, 21.62 )
        };
         \addplot[thick,color=free_speech_aquamarine,mark=diamond*] coordinates {
          (1, 29.65)
          (2, 27.39)
          (3, 25.67)
          (4, 24.42)
          (5, 24.01)
          (6, 24.21)

        }; 

        \addplot[mark size=3.6pt, color=flamingo, mark=*] coordinates {
          (5, 20.35 )
        }; 
         \addplot[mark size=4.2pt, color=free_speech_aquamarine, mark=diamond*] coordinates {
          (6, 24.21)
        };

	\end{groupplot}
\node at ($(group c1r1) + (110pt, 95pt)$) {\ref{grouplegend}};
\end{tikzpicture}

    \caption{Change of validation loss across iterations.}
    \label{fig: iter}
\end{figure}

\paragraph{Effect of Iterative Training}
To show the importance of the iterative training paradigm, we visualize the change of the validation loss in Figure~\ref{fig: iter}~\footnote{Iteration 0 means before the iteration starts but after the initialization stage, equal to a standard seq2seq.}. The forward validation loss is computed as the perplexity of the forward seq2seq on the pseudo context-response pairs obtained from the backward model, vice versa for backward loss. It approximately quantifies the KL divergence between them two~\cite{kim2016sequence,cotterell2018explaining}. As the iteration goes, the knowledge from the backward model is gradually distilled into the forward model. The divergence between them reaches the lowest point at iteration 4, where we stop our model. Table~\ref{table:iter} further displays examples for different iterations. Iteration 0 generates mostly generic responses. Iteration 1 starts to become more diverse but still struggle with fluency and relevance. In the final iteration, it can learn to incorporate novel topics from the non-conversational text yet maintaining the relevance with context.
\begin{table}[!hbtp] \addtolength{\tabcolsep}{-2pt}  
\centering
\begin{tabularx}{1\linewidth}{l|l}
\hline

\multirow{2}{*}{CXT}                  & \makecell[l]{\begin{CJK*}{UTF8}{gbsn}最近又长胖了\end{CJK*}}\\
&Fleshing out again recently.\\
\hline
\multirow{2}{*}{NS}                    & \makecell[l]{ {\begin{CJK*}{UTF8}{gbsn}我也是这样的\end{CJK*} }}\\
& Me too. \\ 
\hline
\multirow{2}{*}{BT}  &  \makecell[l]{\begin{CJK*}{UTF8}{gbsn}哈哈哈莫非已经\error{胖若两人}了\end{CJK*}}\\
& hahaha already \error{as fat as two people}? \\
\Xhline{1pt}

\multirow{2}{*}{CXT}                  & \makecell[l]{\begin{CJK*}{UTF8}{gbsn}爱一个人真的不能跟她表白吗？\end{CJK*}}\\
&Why loving someone but cannot confess?\\
\hline
\multirow{2}{*}{NS}                    & \makecell[l]{ {\begin{CJK*}{UTF8}{gbsn}不一定的\end{CJK*} }}\\
& Not necessarily. \\ 
\hline
\multirow{2}{*}{BT}  &  \makecell[l]{\begin{CJK*}{UTF8}{gbsn}爱一个人\error{不难，难的是}放下一个人。\end{CJK*}}\\
& \error{To} love \error{is easy, to} give up \error{is hard.} \\\hline
\end{tabularx}
\caption{\small Context (CXT), example generations from neucleus sampling (NS) and back-translation (BT). Novel words and syntax patterns are \error{highlighted}.}
\label{table: diverse-example}
\end{table}
\paragraph{Diversity of Generation}
We find the back translation model can generate \emph{both semantically and syntactically} novel responses. Some examples are shown in Table~\ref{table: diverse-example}. To find semantically novel responses, we segment them into phrases and find those containing novel phrases that do not exist on the conversational corpus. As in the first example of Table~\ref{table: diverse-example}, the word \begin{CJK*}{UTF8}{gbsn}胖若两人\end{CJK*} only exists in the non-conversational corpus. The model successfully learnt its semantic meaning and adopt it to generate novel responses. It is also common that the model learns frequent syntax structures from the non-conversational corpus. In the second example, it learnt the pattern of ``To ... is easy, to ... is hard", which appeared frequently in the non-conversational corpus, and utilized it to produce novel responses with the same structure. Note that both generations from the BT model \emph{never appear exactly in the non-conversational corpus}. It must generate them by correctly understanding the meaning of the phrase components instead of memorizing the utterances verbally.
\section{Conclusion and Future Work}
We propose a novel way of diversifying dialogue generation by leveraging non-conversational text. To do so, we collect a large-scale corpus from forum comments, idioms and book snippets. By training the model through iterative back translation, it is able to significantly improve the diversity of generated responses both semantically and syntactically. We compare it with several strong baselines and find it achieved the best overall performance. The model can be potentially improved by filtering the corpus according to different domains, or augmenting with a retrieve-and-rewrite mechanism, which we leave for future work.
\section*{Acknowledgments}
We thank anonymous reviewers for valuable comments. Xiaoyu Shen is supported by IMPRS-CS fellowship. The work is partially funded by DFG collaborative research center SFB 1102.
\bibliography{anthology,acl2020}

\begin{thebibliography}{57}
\expandafter\ifx\csname natexlab\endcsname\relax\def\natexlab#1{#1}\fi

\bibitem[{Artetxe et~al.(2018)Artetxe, Labaka, Agirre, and
  Cho}]{artetxe2017unsupervised}
Mikel Artetxe, Gorka Labaka, Eneko Agirre, and Kyunghyun Cho. 2018.
\newblock Unsupervised neural machine translation.
\newblock \emph{ICLR}.

\bibitem[{Bartl and Spanakis(2017)}]{bartl2017retrieval}
Alexander Bartl and Gerasimos Spanakis. 2017.
\newblock A retrieval-based dialogue system utilizing utterance and context
  embeddings.
\newblock In \emph{2017 16th IEEE International Conference on Machine Learning
  and Applications (ICMLA)}, pages 1120--1125. IEEE.

\bibitem[{Bowman et~al.(2016)Bowman, Vilnis, Vinyals, Dai, Jozefowicz, and
  Bengio}]{bowman2016generating}
Samuel~R Bowman, Luke Vilnis, Oriol Vinyals, Andrew Dai, Rafal Jozefowicz, and
  Samy Bengio. 2016.
\newblock Generating sentences from a continuous space.
\newblock In \emph{Proceedings of The 20th SIGNLL Conference on Computational
  Natural Language Learning}, pages 10--21.

\bibitem[{Chang et~al.(2020)Chang, Adelani, Shen, and
  Demberg}]{chang2020unsupervised}
Ernie Chang, David~Ifeoluwa Adelani, Xiaoyu Shen, and Vera Demberg. 2020.
\newblock Unsupervised pidgin text generation by pivoting english data and
  self-training.
\newblock \emph{arXiv preprint arXiv:2003.08272}.

\bibitem[{Cotterell and Kreutzer(2018)}]{cotterell2018explaining}
Ryan Cotterell and Julia Kreutzer. 2018.
\newblock Explaining and generalizing back-translation through wake-sleep.
\newblock \emph{arXiv preprint arXiv:1806.04402}.

\bibitem[{Cs{\'a}ky et~al.(2019)Cs{\'a}ky, Purgai, and
  Recski}]{csaky-etal-2019-improving}
Rich{\'a}rd Cs{\'a}ky, Patrik Purgai, and G{\'a}bor Recski. 2019.
\newblock \href {https://doi.org/10.18653/v1/P19-1567} {Improving neural
  conversational models with entropy-based data filtering}.
\newblock In \emph{Proceedings of the 57th Annual Meeting of the Association
  for Computational Linguistics}, pages 5650--5669, Florence, Italy.
  Association for Computational Linguistics.

\bibitem[{Devlin et~al.(2019)Devlin, Chang, Lee, and
  Toutanova}]{devlin2019bert}
Jacob Devlin, Ming-Wei Chang, Kenton Lee, and Kristina Toutanova. 2019.
\newblock Bert: Pre-training of deep bidirectional transformers for language
  understanding.
\newblock In \emph{Proceedings of the 2019 Conference of the North American
  Chapter of the Association for Computational Linguistics: Human Language
  Technologies, Volume 1 (Long and Short Papers)}, pages 4171--4186.

\bibitem[{Dinan et~al.(2019)Dinan, Roller, Shuster, Fan, Auli, and
  Weston}]{dinan2018wizard}
Emily Dinan, Stephen Roller, Kurt Shuster, Angela Fan, Michael Auli, and Jason
  Weston. 2019.
\newblock Wizard of wikipedia: Knowledge-powered conversational agents.
\newblock \emph{ICLR}.

\bibitem[{Gao et~al.(2019)Gao, Zhang, Lee, Galley, Brockett, Gao, and
  Dolan}]{gao2019structuring}
Xiang Gao, Yizhe Zhang, Sungjin Lee, Michel Galley, Chris Brockett, Jianfeng
  Gao, and Bill Dolan. 2019.
\newblock Structuring latent spaces for stylized response generation.
\newblock In \emph{Proceedings of the 2019 Conference on Empirical Methods in
  Natural Language Processing and the 9th International Joint Conference on
  Natural Language Processing (EMNLP-IJCNLP)}, pages 1814--1823.

\bibitem[{Ghazvininejad et~al.(2018)Ghazvininejad, Brockett, Chang, Dolan, Gao,
  Yih, and Galley}]{ghazvininejad2018knowledge}
Marjan Ghazvininejad, Chris Brockett, Ming-Wei Chang, Bill Dolan, Jianfeng Gao,
  Wen-tau Yih, and Michel Galley. 2018.
\newblock A knowledge-grounded neural conversation model.
\newblock In \emph{Thirty-Second AAAI Conference on Artificial Intelligence}.

\bibitem[{Hoang et~al.(2018)Hoang, Koehn, Haffari, and
  Cohn}]{hoang2018iterative}
Vu~Cong~Duy Hoang, Philipp Koehn, Gholamreza Haffari, and Trevor Cohn. 2018.
\newblock Iterative back-translation for neural machine translation.
\newblock In \emph{Proceedings of the 2nd Workshop on Neural Machine
  Translation and Generation}, pages 18--24.

\bibitem[{Hochreiter and Schmidhuber(1997)}]{hochreiter1997long}
Sepp Hochreiter and Jurgen Schmidhuber. 1997.
\newblock Long short-term memory.
\newblock \emph{Neural Computation}, 9(8):1735--1780.

\bibitem[{Holtzman et~al.(2019)Holtzman, Buys, Forbes, and
  Choi}]{holtzman2019curious}
Ari Holtzman, Jan Buys, Maxwell Forbes, and Yejin Choi. 2019.
\newblock The curious case of neural text degeneration.
\newblock \emph{arXiv preprint arXiv:1904.09751}.

\bibitem[{Kim and Rush(2016)}]{kim2016sequence}
Yoon Kim and Alexander~M Rush. 2016.
\newblock Sequence-level knowledge distillation.
\newblock In \emph{Proceedings of the 2016 Conference on Empirical Methods in
  Natural Language Processing}, pages 1317--1327.

\bibitem[{Kingma and Ba(2015)}]{kingma2014adam}
Diederik Kingma and Jimmy Ba. 2015.
\newblock Adam: A method for stochastic optimization.
\newblock \emph{ICLR}.

\bibitem[{Klein et~al.(2017)Klein, Kim, Deng, Senellart, and
  Rush}]{klein2017opennmt}
Guillaume Klein, Yoon Kim, Yuntian Deng, Jean Senellart, and Alexander Rush.
  2017.
\newblock Opennmt: Open-source toolkit for neural machine translation.
\newblock In \emph{Proceedings of ACL 2017, System Demonstrations}, pages
  67--72.

\bibitem[{Koehn and Schroeder(2007)}]{koehn2007experiments}
Philipp Koehn and Josh Schroeder. 2007.
\newblock Experiments in domain adaptation for statistical machine translation.
\newblock In \emph{Proceedings of the second workshop on statistical machine
  translation}, pages 224--227.

\bibitem[{Lample et~al.(2018{\natexlab{a}})Lample, Conneau, Denoyer, and
  Ranzato}]{lample2017unsupervised}
Guillaume Lample, Alexis Conneau, Ludovic Denoyer, and Marc'Aurelio Ranzato.
  2018{\natexlab{a}}.
\newblock Unsupervised machine translation using monolingual corpora only.
\newblock \emph{ICLR}.

\bibitem[{Lample et~al.(2018{\natexlab{b}})Lample, Ott, Conneau, Denoyer
  et~al.}]{lample2018phrase}
Guillaume Lample, Myle Ott, Alexis Conneau, Ludovic Denoyer, et~al.
  2018{\natexlab{b}}.
\newblock Phrase-based \& neural unsupervised machine translation.
\newblock In \emph{Proceedings of the 2018 Conference on Empirical Methods in
  Natural Language Processing}, pages 5039--5049.

\bibitem[{Li et~al.(2016{\natexlab{a}})Li, Galley, Brockett, Gao, and
  Dolan}]{li2015diversity}
Jiwei Li, Michel Galley, Chris Brockett, Jianfeng Gao, and Bill Dolan.
  2016{\natexlab{a}}.
\newblock A diversity-promoting objective function for neural conversation
  models.
\newblock In \emph{Proceedings of the 2016 Conference of the North American
  Chapter of the Association for Computational Linguistics: Human Language
  Technologies}, pages 110--119.

\bibitem[{Li et~al.(2016{\natexlab{b}})Li, Galley, Brockett, Spithourakis, Gao,
  and Dolan}]{li2016persona}
Jiwei Li, Michel Galley, Chris Brockett, Georgios~P Spithourakis, Jianfeng Gao,
  and Bill Dolan. 2016{\natexlab{b}}.
\newblock A persona-based neural conversation model.
\newblock \emph{arXiv preprint arXiv:1603.06155}.

\bibitem[{Li and Jurafsky(2017)}]{li2016neural}
Jiwei Li and Dan Jurafsky. 2017.
\newblock Neural net models for open-domain discourse coherence.
\newblock \emph{EMNLP}.

\bibitem[{Li et~al.(2016{\natexlab{c}})Li, Monroe, and Jurafsky}]{li2016simple}
Jiwei Li, Will Monroe, and Dan Jurafsky. 2016{\natexlab{c}}.
\newblock \href {http://arxiv.org/abs/1611.08562} {A simple, fast diverse
  decoding algorithm for neural generation}.
\newblock \emph{CoRR}, abs/1611.08562.

\bibitem[{Li et~al.(2016{\natexlab{d}})Li, Monroe, Ritter, Jurafsky, Galley,
  and Gao}]{li2016deep}
Jiwei Li, Will Monroe, Alan Ritter, Dan Jurafsky, Michel Galley, and Jianfeng
  Gao. 2016{\natexlab{d}}.
\newblock Deep reinforcement learning for dialogue generation.
\newblock In \emph{Proceedings of the 2016 Conference on Empirical Methods in
  Natural Language Processing}, pages 1192--1202.

\bibitem[{Li et~al.(2017)Li, Monroe, Shi, Jean, Ritter, and
  Jurafsky}]{li2017adversarial}
Jiwei Li, Will Monroe, Tianlin Shi, S{\.e}bastien Jean, Alan Ritter, and Dan
  Jurafsky. 2017.
\newblock Adversarial learning for neural dialogue generation.
\newblock In \emph{Proceedings of the 2017 Conference on Empirical Methods in
  Natural Language Processing}, pages 2157--2169.

\bibitem[{Lu et~al.(2017)Lu, Kannan, Yang, Parikh, and Batra}]{lu2017best}
Jiasen Lu, Anitha Kannan, Jianwei Yang, Devi Parikh, and Dhruv Batra. 2017.
\newblock Best of both worlds: Transferring knowledge from discriminative
  learning to a generative visual dialog model.
\newblock In \emph{Advances in Neural Information Processing Systems}, pages
  314--324.

\bibitem[{Luan et~al.(2017)Luan, Brockett, Dolan, Gao, and
  Galley}]{luan2017multi}
Yi~Luan, Chris Brockett, Bill Dolan, Jianfeng Gao, and Michel Galley. 2017.
\newblock Multi-task learning for speaker-role adaptation in neural
  conversation models.
\newblock In \emph{Proceedings of the Eighth International Joint Conference on
  Natural Language Processing (Volume 1: Long Papers)}, pages 605--614.

\bibitem[{Luo et~al.(2018)Luo, Price, Cohen, and
  Shakhnarovich}]{luo2018discriminability}
Ruotian Luo, Brian Price, Scott Cohen, and Gregory Shakhnarovich. 2018.
\newblock Discriminability objective for training descriptive captions.
\newblock In \emph{Proceedings of the IEEE Conference on Computer Vision and
  Pattern Recognition}, pages 6964--6974.

\bibitem[{Luong et~al.(2016)Luong, Le, Sutskever, Vinyals, and
  Kaiser}]{luong2015multi}
Minh-Thang Luong, Quoc~V Le, Ilya Sutskever, Oriol Vinyals, and Lukasz Kaiser.
  2016.
\newblock Multi-task sequence to sequence learning.
\newblock \emph{ICLR}.

\bibitem[{May et~al.(2019)May, Wang, Bordia, Bowman, and
  Rudinger}]{may2019measuring}
Chandler May, Alex Wang, Shikha Bordia, Samuel Bowman, and Rachel Rudinger.
  2019.
\newblock On measuring social biases in sentence encoders.
\newblock In \emph{Proceedings of the 2019 Conference of the North American
  Chapter of the Association for Computational Linguistics: Human Language
  Technologies, Volume 1 (Long and Short Papers)}, pages 622--628.

\bibitem[{Niu and Bansal(2018)}]{niu2018polite}
Tong Niu and Mohit Bansal. 2018.
\newblock Polite dialogue generation without parallel data.
\newblock \emph{Transactions of the Association for Computational Linguistics},
  6:373--389.

\bibitem[{Papineni et~al.(2002)Papineni, Roukos, Ward, and
  Zhu}]{papineni2002bleu}
Kishore Papineni, Salim Roukos, Todd Ward, and Wei-Jing Zhu. 2002.
\newblock Bleu: a method for automatic evaluation of machine translation.
\newblock In \emph{Proceedings of the 40th annual meeting on association for
  computational linguistics}, pages 311--318. Association for Computational
  Linguistics.

\bibitem[{Sennrich et~al.(2016)Sennrich, Haddow, and
  Birch}]{sennrich2016improving}
Rico Sennrich, Barry Haddow, and Alexandra Birch. 2016.
\newblock Improving neural machine translation models with monolingual data.
\newblock In \emph{Proceedings of the 54th Annual Meeting of the Association
  for Computational Linguistics (Volume 1: Long Papers)}, pages 86--96.

\bibitem[{Serban et~al.(2017{\natexlab{a}})Serban, Sankar, Germain, Zhang, Lin,
  Subramanian, Kim, Pieper, Chandar, Ke et~al.}]{serban2017deep}
Iulian~V Serban, Chinnadhurai Sankar, Mathieu Germain, Saizheng Zhang, Zhouhan
  Lin, Sandeep Subramanian, Taesup Kim, Michael Pieper, Sarath Chandar,
  Nan~Rosemary Ke, et~al. 2017{\natexlab{a}}.
\newblock A deep reinforcement learning chatbot.
\newblock \emph{arXiv preprint arXiv:1709.02349}.

\bibitem[{Serban et~al.(2017{\natexlab{b}})Serban, Sordoni, Lowe, Charlin,
  Pineau, Courville, and Bengio}]{serban2016hierarchical}
Iulian~Vlad Serban, Alessandro Sordoni, Ryan Lowe, Laurent Charlin, Joelle
  Pineau, Aaron Courville, and Yoshua Bengio. 2017{\natexlab{b}}.
\newblock A hierarchical latent variable encoder-decoder model for generating
  dialogues.
\newblock In \emph{Thirty-First AAAI Conference on Artificial Intelligence},
  pages 3295--3301.

\bibitem[{Shang et~al.(2015{\natexlab{a}})Shang, Lu, and Li}]{shang2015neural}
Lifeng Shang, Zhengdong Lu, and Hang Li. 2015{\natexlab{a}}.
\newblock Neural responding machine for short-text conversation.
\newblock \emph{arXiv preprint arXiv:1503.02364}.

\bibitem[{Shang et~al.(2015{\natexlab{b}})Shang, Lu, and
  Li}]{shang-etal-2015-neural}
Lifeng Shang, Zhengdong Lu, and Hang Li. 2015{\natexlab{b}}.
\newblock \href {https://doi.org/10.3115/v1/P15-1152} {Neural responding
  machine for short-text conversation}.
\newblock In \emph{Proceedings of the 53rd Annual Meeting of the Association
  for Computational Linguistics and the 7th International Joint Conference on
  Natural Language Processing (Volume 1: Long Papers)}, pages 1577--1586,
  Beijing, China. Association for Computational Linguistics.

\bibitem[{Shen et~al.(2017{\natexlab{a}})Shen, Oualil, Greenberg, Singh, and
  Klakow}]{shen2017estimation}
Xiaoyu Shen, Youssef Oualil, Clayton Greenberg, Mittul Singh, and Dietrich
  Klakow. 2017{\natexlab{a}}.
\newblock Estimation of gap between current language models and human
  performance.
\newblock \emph{Proc. Interspeech 2017}, pages 553--557.

\bibitem[{Shen et~al.(2018{\natexlab{a}})Shen, Su, Li, and
  Klakow}]{shen2018nexus}
Xiaoyu Shen, Hui Su, Wenjie Li, and Dietrich Klakow. 2018{\natexlab{a}}.
\newblock Nexus network: Connecting the preceding and the following in dialogue
  generation.
\newblock In \emph{Proceedings of the 2018 Conference on Empirical Methods in
  Natural Language Processing}, pages 4316--4327.

\bibitem[{Shen et~al.(2017{\natexlab{b}})Shen, Su, Li, Li, Niu, Zhao, Aizawa,
  and Long}]{shen2017conditional}
Xiaoyu Shen, Hui Su, Yanran Li, Wenjie Li, Shuzi Niu, Yang Zhao, Akiko Aizawa,
  and Guoping Long. 2017{\natexlab{b}}.
\newblock A conditional variational framework for dialog generation.
\newblock In \emph{Proceedings of the 55th Annual Meeting of the Association
  for Computational Linguistics (Volume 2: Short Papers)}, volume~2, pages
  504--509.

\bibitem[{Shen et~al.(2018{\natexlab{b}})Shen, Su, Niu, and
  Demberg}]{shen2018improving}
Xiaoyu Shen, Hui Su, Shuzi Niu, and Vera Demberg. 2018{\natexlab{b}}.
\newblock Improving variational encoder-decoders in dialogue generation.
\newblock \emph{AAAI}, pages 5456--5463.

\bibitem[{Shen et~al.(2019{\natexlab{a}})Shen, Suzuki, Inui, Su, Klakow, and
  Sekine}]{shen2019select}
Xiaoyu Shen, Jun Suzuki, Kentaro Inui, Hui Su, Dietrich Klakow, and Satoshi
  Sekine. 2019{\natexlab{a}}.
\newblock Select and attend: Towards controllable content selection in text
  generation.
\newblock \emph{arXiv preprint arXiv:1909.04453}.

\bibitem[{Shen et~al.(2019{\natexlab{b}})Shen, Zhao, Su, and
  Klakow}]{shen2019improving}
Xiaoyu Shen, Yang Zhao, Hui Su, and Dietrich Klakow. 2019{\natexlab{b}}.
\newblock Improving latent alignment in text summarization by generalizing the
  pointer generator.
\newblock In \emph{Proceedings of the 2019 Conference on Empirical Methods in
  Natural Language Processing and the 9th International Joint Conference on
  Natural Language Processing (EMNLP-IJCNLP)}, pages 3753--3764.

\bibitem[{Su et~al.(2019{\natexlab{a}})Su, Hsu, Tuan, and
  Lee}]{su2019personalized}
Feng-Guang Su, Aliyah~R Hsu, Yi-Lin Tuan, and Hung-Yi Lee. 2019{\natexlab{a}}.
\newblock Personalized dialogue response generation learned from monologues.
\newblock \emph{Proc. Interspeech 2019}, pages 4160--4164.

\bibitem[{Su et~al.(2018)Su, Shen, Hu, Li, and Chen}]{su2018dialogue}
Hui Su, Xiaoyu Shen, Pengwei Hu, Wenjie Li, and Yun Chen. 2018.
\newblock Dialogue generation with gan.
\newblock In \emph{Thirty-Second AAAI Conference on Artificial Intelligence}.

\bibitem[{Su et~al.(2019{\natexlab{b}})Su, Shen, Zhang, Sun, Hu, Niu, and
  Zhou}]{su2019improving}
Hui Su, Xiaoyu Shen, Rongzhi Zhang, Fei Sun, Pengwei Hu, Cheng Niu, and Jie
  Zhou. 2019{\natexlab{b}}.
\newblock Improving multi-turn dialogue modelling with utterance rewriter.
\newblock \emph{arXiv preprint arXiv:1906.07004}.

\bibitem[{Subramanian et~al.(2019)Subramanian, Lample, Smith, Denoyer, Ranzato,
  and Boureau}]{subramanian2018multiple}
Sandeep Subramanian, Guillaume Lample, Eric~Michael Smith, Ludovic Denoyer,
  Marc'Aurelio Ranzato, and Y-Lan Boureau. 2019.
\newblock Multiple-attribute text style transfer.
\newblock \emph{ICLR}.

\bibitem[{Vijayakumar et~al.(2018)Vijayakumar, Cogswell, Selvaraju, Sun, Lee,
  Crandall, and Batra}]{vijayakumar2016diverse}
Ashwin~K Vijayakumar, Michael Cogswell, Ramprasaath~R Selvaraju, Qing Sun,
  Stefan Lee, David~J Crandall, and Dhruv Batra. 2018.
\newblock Diverse beam search for improved description of complex scenes.
\newblock \emph{AAAI}, pages 7371--7379.

\bibitem[{Vinyals and Le(2015)}]{vinyals2015neural}
Oriol Vinyals and Quoc~V. Le. 2015.
\newblock \href {http://arxiv.org/abs/1506.05869} {A neural conversational
  model}.
\newblock \emph{CoRR}, abs/1506.05869.

\bibitem[{Wang et~al.(2017)Wang, Jojic, Brockett, and
  Nyberg}]{wang2017steering}
Di~Wang, Nebojsa Jojic, Chris Brockett, and Eric Nyberg. 2017.
\newblock Steering output style and topic in neural response generation.
\newblock In \emph{Proceedings of the 2017 Conference on Empirical Methods in
  Natural Language Processing}, pages 2140--2150.

\bibitem[{Wu et~al.(2017)Wu, Wu, Xing, Zhou, and Li}]{wu2017sequential}
Yu~Wu, Wei Wu, Chen Xing, Ming Zhou, and Zhoujun Li. 2017.
\newblock Sequential matching network: A new architecture for multi-turn
  response selection in retrieval-based chatbots.
\newblock In \emph{Proceedings of the 55th Annual Meeting of the Association
  for Computational Linguistics (Volume 1: Long Papers)}, pages 496--505.

\bibitem[{Zhang et~al.(2018{\natexlab{a}})Zhang, Dinan, Urbanek, Szlam, Kiela,
  and Weston}]{zhang2018personalizing}
Saizheng Zhang, Emily Dinan, Jack Urbanek, Arthur Szlam, Douwe Kiela, and Jason
  Weston. 2018{\natexlab{a}}.
\newblock Personalizing dialogue agents: I have a dog, do you have pets too?
\newblock In \emph{Proceedings of the 56th Annual Meeting of the Association
  for Computational Linguistics (Volume 1: Long Papers)}, pages 2204--2213.

\bibitem[{Zhang et~al.(2018{\natexlab{b}})Zhang, Galley, Gao, Gan, Li,
  Brockett, and Dolan}]{zhang2018generating}
Yizhe Zhang, Michel Galley, Jianfeng Gao, Zhe Gan, Xiujun Li, Chris Brockett,
  and Bill Dolan. 2018{\natexlab{b}}.
\newblock Generating informative and diverse conversational responses via
  adversarial information maximization.
\newblock In \emph{Advances in Neural Information Processing Systems}, pages
  1810--1820.

\bibitem[{Zhao et~al.(2017)Zhao, Zhao, and Eskenazi}]{zhao2017learning}
Tiancheng Zhao, Ran Zhao, and Maxine Eskenazi. 2017.
\newblock Learning discourse-level diversity for neural dialog models using
  conditional variational autoencoders.
\newblock In \emph{Proceedings of the 55th Annual Meeting of the Association
  for Computational Linguistics (Volume 1: Long Papers)}, volume~1, pages
  654--664.

\bibitem[{Zhao et~al.(2019)Zhao, Shen, Bi, and Aizawa}]{zhao2019unsupervised}
Yang Zhao, Xiaoyu Shen, Wei Bi, and Akiko Aizawa. 2019.
\newblock Unsupervised rewriter for multi-sentence compression.
\newblock In \emph{Proceedings of the 57th Annual Meeting of the Association
  for Computational Linguistics}, pages 2235--2240.

\bibitem[{Zhao et~al.(2018)Zhao, Shen, Senuma, and
  Aizawa}]{zhao2018comprehensive}
Yang Zhao, Xiaoyu Shen, Hajime Senuma, and Akiko Aizawa. 2018.
\newblock A comprehensive study: Sentence compression with linguistic
  knowledge-enhanced gated neural network.
\newblock \emph{Data \& Knowledge Engineering}, 117:307--318.

\bibitem[{Zhou et~al.(2018)Zhou, Huang, Zhang, Zhu, and
  Liu}]{zhou2018emotional}
Hao Zhou, Minlie Huang, Tianyang Zhang, Xiaoyan Zhu, and Bing Liu. 2018.
\newblock Emotional chatting machine: Emotional conversation generation with
  internal and external memory.
\newblock In \emph{Thirty-Second AAAI Conference on Artificial Intelligence}.

\end{thebibliography}
\bibliographystyle{acl_natbib}

\end{document}